\documentclass[11pt]{article}

\usepackage[margin=1in]{geometry}
\usepackage[T1]{fontenc}
\usepackage[utf8]{inputenc}
\usepackage{lmodern}
\usepackage{microtype}
\usepackage{graphicx}
\usepackage{booktabs}
\usepackage{hyperref}
\usepackage{xcolor}
\usepackage{amsmath}
\usepackage{natbib}
\usepackage{authblk}
\usepackage{titlesec}
\usepackage{longtable}

\hypersetup{colorlinks=true, linkcolor=blue!50!black, citecolor=blue!50!black,
            urlcolor=blue!50!black}

\titleformat{\section}{\normalfont\large\bfseries}{\thesection}{0.6em}{}
\titleformat{\subsection}{\normalfont\normalsize\bfseries}{\thesubsection}{0.6em}{}

\title{\textbf{12 Angry AI Agents:\\
Evaluating Multi-Agent LLM Decision-Making\\
Through Cinematic Jury Deliberation}}
\author{\textbf{Ahmet Bahaddin Ersoz}\thanks{ORCID: \href{https://orcid.org/0000-0001-6297-7501}{0000-0001-6297-7501}}}
\affil{\url{https://ahmetbersoz.github.io/12-angry-ai-agents/}}

\date{}

\begin{document}
\maketitle

\begin{abstract}
  \noindent
  What if the twelve jurors of Sidney Lumet's 12 Angry Men (1957) were not men, but large language models? Would the one juror who disagrees still be able to change everyone's mind? This paper instantiates that scenario as a multi-agent benchmark for LLM deliberation: twelve agents, each conditioned on a film-faithful persona, debate the film's murder case using multi-agent framework.
  Two models representing opposite ends of the RLHF spectrum are tested: GPT-4o (closed-source, heavy alignment) and Llama-4-Scout (open-weight, lighter alignment), across three conditions (baseline, open-minded prompt, no initial vote), with $N = 3$ replications per cell (18 runs total).
  Three findings emerge.
  (i) Seventeen of eighteen runs end in a hung jury (a state where the jury fails to reach a unanimous verdict); the film's central event, gradual minority-to-majority persuasion, almost never occurs, indicating that anchoring is the dominant failure mode of current LLMs in this setting.
  (ii) The two models exhibit sharply different internal dynamics: GPT-4o produces a mean of 1.0 vote changes per run across all conditions, while Llama-4-Scout ranges from 2.0 (baseline) to 6.0 (open-minded prompt), and is the only model to reach a NOT\_GUILTY verdict (1 of 3 runs in the no-initial-vote condition). The same ``open-minded'' instruction is internalized by Llama and ignored by GPT-4o.
  (iii) This asymmetry suggests that the intensity of RLHF alignment training, not model capability, is the primary determinant of deliberative flexibility in multi-agent settings. Flexibility, not capability, tracks human deliberation.
  The work is framed as an exploratory study and discusses implications for jury-of-LLMs evaluation and multi-agent debate.
\end{abstract}

\section{Introduction}
\label{sec:intro}

Large language models are increasingly composed into multi-agent systems where several model instances exchange messages to reach a decision. The architecture appears across many recent applications: panels of LLM judges that evaluate model outputs, multi-agent debate frameworks that aim to improve reasoning by surfacing disagreement \citep{du2023debate,liang2023encouraging}, code-review and pair-programming systems with specialized agent roles \citep{qian2023chatdev}, and simulated councils used in policy and design settings \citep{park2023generative}. The premise common to all of these is that deliberation between agents (the back-and-forth weighing of evidence and updating of positions) produces better outcomes than a single model asked the same question.

This premise rests on an empirical claim that has not been carefully tested: that LLM agents, when placed in a deliberative setting, actually deliberate. They might instead anchor on whatever position they begin with, recite arguments without integrating counter-arguments, and produce the appearance of deliberation without its substance. If so, the mechanism that justifies multi-agent systems is largely cosmetic, and the resulting verdicts are determined by initial conditions rather than by argument quality.

To probe this question, a deliberative scenario with two properties is needed: (i) a known ground-truth trajectory of mind-changing, against which agent behavior can be compared; and (ii) richly defined, individuated participants whose disagreements have a recognizable shape. Sidney Lumet's 1957 film 12 Angry Men provides both \citep{lumet1957angry}. Twelve jurors begin a murder deliberation 11--1 in favor of GUILTY, and over the course of roughly ninety minutes the lone NOT\_GUILTY dissenter (Juror\_8) persuades the other eleven to acquit. The order in which jurors flip is fixed and well documented, the personality and bias of each juror is film-canonical, and the case file is small and self-contained.

This paper instantiates that scenario as a multi-agent LLM benchmark. Twelve agents, each conditioned on a film-faithful persona, deliberate the same case using Microsoft AutoGen's \texttt{SelectorGroupChat} pattern \citep{wu2023autogen}. The experimental design tests two models that represent opposite ends of the alignment spectrum in depth: GPT-4o, the flagship of the proprietary, closed-source ecosystem with one of the heaviest RLHF (reinforcement learning from human feedback) pipelines in the industry; and Llama-4-Scout, the flagship of the open-weight ecosystem with a publicly documented, lighter alignment stack. This pairing is deliberate: it tests whether the dominant development philosophies of closed-source safety-first alignment and open-source flexibility-first alignment produce different deliberative behaviors under identical conditions.

The findings are sharp. Seventeen of eighteen runs end in a hung jury. GPT-4o produces essentially flat deliberations (mean 1.0 vote changes per run) regardless of prompt condition, while Llama-4-Scout's vote-change rate triples from 2.0 (baseline) to 6.0 when given an ``open-minded'' prompt instruction. In the no-initial-vote condition, Llama is the only model to reach a NOT\_GUILTY verdict (1 of 3 runs). The general assumption in the AI literature that ``bigger is better'' and that model capability uniformly predicts task performance is inverted here: in multi-agent deliberation, flexibility (not capability) is what tracks human-like persuasion dynamics.

The work is framed as an exploratory study: each cell has $N = 3$, and directional findings are reported rather than significance tests. The contributions are:
\begin{enumerate}
  \item A film-grounded benchmark for multi-agent LLM deliberation, with a well-known ground-truth cascade and twelve canonical personas.
  \item A controlled comparison of two RLHF paradigms (heavy vs.\ light alignment) on the same deliberative task, providing evidence that RLHF intensity, not model size, governs deliberative rigidity.
  \item Prompt-ablation and vote-conditioning ablation evidence that anchoring is the dominant failure mode, and that the same de-anchoring intervention works asymmetrically across alignment regimes.
  \item A qualitative comparison between film deliberation and LLM deliberation, identifying the specific surface features LLMs reproduce and the specific mechanisms they fail to reproduce.
\end{enumerate}

\section{Related Work}
\label{sec:related}

\paragraph{Multi-agent LLM frameworks.} A wave of recent systems compose multiple LLM instances into deliberative or collaborative pipelines. AutoGen \citep{wu2023autogen} provides the orchestration substrate used in this paper. CAMEL \citep{li2023camel} instantiates two agents (a ``user'' and an ``assistant'') that role-play through a task. ChatDev \citep{qian2023chatdev} assembles software-engineering teams of LLM agents with specialized roles. Generative Agents \citep{park2023generative} populates a simulated town with persona-conditioned LLM agents whose interactions produce emergent social behavior. This paper adopts the AutoGen substrate but focuses on the quality of deliberation between agents rather than on system architecture.

\paragraph{Multi-agent debate and reasoning.} \citet{du2023debate} and \citet{liang2023encouraging} propose multi-agent debate as a route to improved factuality and reasoning, on the premise that disagreement between agents surfaces errors that a single agent would miss. The present results raise a complementary concern: in scenarios that require one agent to persuade another, current LLMs may produce parallel monologues rather than genuine persuasion, undermining the assumed mechanism.

\paragraph{Anchoring and cognitive biases in LLMs.} The anchoring effect (disproportionate weight on initially presented information) is a long-studied human bias \citep{tversky1974anchoring}. Recent work has documented LLM-side analogues \citep{suri2024anchoring,echterhoff2024cognitive}. This study extends the literature to a multi-turn social setting: anchoring not to a presented number, but to a position taken under persona conditioning.

\paragraph{Persona conditioning and consistency.} A growing body of work studies how stably LLMs maintain assigned personas \citep{park2023generative,wang2024persona}. High stylistic fidelity is observed in the present experiments, but persona consistency may itself contribute to the anchoring effect; once an agent has spoken ``in character'' for several turns, it may be reluctant to adopt a position out of character.

\paragraph{LLM panels as judges.} A practical motivation for this work is the increasing use of LLM panels (juries) to evaluate other LLM outputs \citep{zheng2023judging}. If panel members anchor on initial impressions and rarely revise them under cross-examination, the value-add of using a panel rather than a single judge is reduced.

\paragraph{Narrative as benchmark.} Using a known narrative cascade as ground truth for multi-agent behavior is, to the author's knowledge, novel. The closest precedent is the use of board games and structured dialogues (e.g., Diplomacy in \citealt{bakhtin2022diplomacy}) as testbeds for multi-agent strategy and negotiation. 12 Angry Men differs in that the success criterion is persuasion of the majority by a minority, which is precisely the dynamic absent from current multi-agent LLM systems.

\section{Methodology}
\label{sec:methods}

\subsection{Case and Personas}
The deliberative scenario of Sidney Lumet's 12 Angry Men (1957) is instantiated as follows. Twelve juror agents debate a first-degree murder case in which a young defendant is accused of stabbing his father. The case file is encoded as eight discrete evidence items: the switchblade knife, the old man's eyewitness testimony, the woman across the street's testimony, the El train timing, the alibi, the threat (``I'll kill you''), the stab-wound angle, and the old man's limp. Each agent has access to the full case file at the start of deliberation. The complete case file and evidence descriptions are reproduced in Appendix~\ref{app:kb}.

Each of the twelve jurors is a film-faithful persona that specifies an occupation, a set of biases, emotional triggers, a speaking style, an initial vote, and a list of key arguments the juror is expected to raise. The initial vote distribution is eleven GUILTY and one NOT\_GUILTY (Juror\_8 is the lone dissenter), mirroring the film's opening tableau. Full persona definitions are provided in Appendix~\ref{app:personas}.

\subsection{Model Selection}
\label{sec:modelselection}

Rather than surveying a broad set of model families at shallow depth, this study deliberately selects two models that represent the dominant paradigms of the current LLM landscape:

GPT-4o represents the proprietary, closed-source paradigm. It has undergone one of the most intensive RLHF pipelines in the industry, producing a model that is highly ``safe,'' stylistically consistent, and resistant to deviating from established positions. It serves as the flagship of the safety-first, closed-box development philosophy.

Llama-4-Scout represents the open-weight paradigm. Its RLHF stack is publicly documented and known to be lighter than those of proprietary competitors. It is more open-ended in character embodiment (role-play) and more responsive to in-context steering. It serves as the flagship of the flexibility-first, open-source development philosophy.

This pairing is not arbitrary. By testing the two dominant development philosophies under identical deliberative conditions, the findings are applicable to practitioners working in either ecosystem. The question is not ``which model is better?'' but ``how does alignment philosophy shape deliberative behavior?''

\subsection{Deliberation Engine}
The deliberation is orchestrated by Microsoft AutoGen's \texttt{SelectorGroupChat} pattern, in which a model-driven selector chooses which agent speaks next at each turn. Speaker selection encodes turn-rotation rules and a dissent-priority heuristic that ensures Juror\_8 receives speaking opportunities even when in the minority.

A structured-output vote prompt is issued every twelve turns, requiring each juror to declare their current vote (GUILTY / NOT\_GUILTY). The deliberation terminates when (a) all twelve jurors have converged on a single vote (yielding a verdict of GUILTY or NOT\_GUILTY), or (b) an early-stopping criterion is met, in which case the verdict is recorded as HUNG\_JURY.

\paragraph{Early stopping.} To avoid wasteful computation in deliberations that have clearly stalled, an early-stopping mechanism is employed: if three consecutive voting rounds (e.g., turns 36, 48, and 60) produce zero vote changes across all twelve jurors, the deliberation is terminated and the result is recorded as HUNG\_JURY. The maximum turn budget is set to \texttt{max\_turns}~$= 150$, but early stopping typically terminates runs well before this limit (observed range: 36--84 turns; see Figure~\ref{fig:earlystop}). This mechanism ensures that the HUNG\_JURY classification reflects genuine deliberative stagnation rather than mere time exhaustion.

\subsection{Experimental Conditions}
Six configurations are tested: two models crossed with three prompt/conditioning conditions. Each cell is replicated three times, yielding 18 runs total. Configurations are summarized in Table~\ref{tab:conditions}.

\begin{table}[t]
  \centering
  \caption{Experimental conditions. All runs use $t = 0.9$, \texttt{max\_turns} $= 150$ with early stopping. $N = 3$ per cell, 18 runs total.}
  \label{tab:conditions}
  \small
  \begin{tabular}{llp{8cm}}
    \toprule
    Condition          & Models                & Description                                                                                                                          \\
    \midrule
    Baseline           & GPT-4o, Llama-4-Scout & Standard persona prompts with 11 GUILTY / 1 NOT\_GUILTY initial vote conditioning                                                    \\
    No initial vote    & GPT-4o, Llama-4-Scout & Initial vote removed from persona; jurors decide from scratch based on evidence                                                      \\
    Open-minded prompt & GPT-4o, Llama-4-Scout & Adds rule: ``Weigh ALL evidence fairly. Do not cling to your first instinct; if a counter-argument is sound, update your position.'' \\
    \bottomrule
  \end{tabular}
\end{table}

The study is exploratory: each cell has $N = 3$, and directional findings are reported rather than statistical significance tests.

\subsection{Metrics}
For each run the following are recorded: verdict ($\in \{$GUILTY, NOT\_GUILTY, HUNG\_JURY$\}$); total turns to termination; vote-change timeline (an ordered list of (turn, juror, from-vote, to-vote) tuples); number of vote changes and cascade velocity (changes per turn); first-flip order (the sequence in which jurors flip GUILTY $\rightarrow$ NOT\_GUILTY for the first time); and Spearman~$\rho$ (rank correlation between observed first-flip order and the film's canonical order, computed only when $\geq 2$ jurors flip).

\section{Results}
\label{sec:results}

All numeric claims below trace to per-run results. Across 18 runs in 6 configurations, the pattern is clear: anchoring dominates, but its grip varies sharply by model.

\subsection{Most Runs End in a Hung Jury}
\begin{figure}[t]
  \centering
  \includegraphics[width=0.85\linewidth]{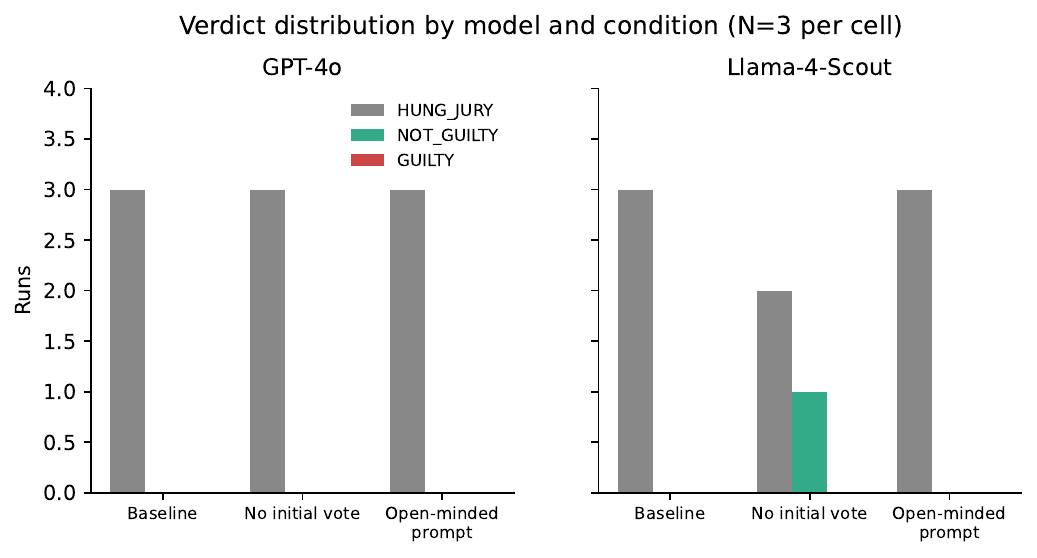}
  \caption{Verdict distribution by model and condition ($N = 3$ per cell). GPT-4o produces hung juries in all 9 runs. Llama-4-Scout produces 1 NOT\_GUILTY verdict in the no-initial-vote condition.}
  \label{fig:verdicts}
\end{figure}

Across all configurations, 17 of 18 runs end in HUNG\_JURY. The single exception is Llama-4-Scout in the no-initial-vote condition (run 2 of 3), which converges to a NOT\_GUILTY verdict in 64 turns with 6 vote changes. Figure~\ref{fig:verdicts} shows the verdict distribution by model and condition.

GPT-4o produces a hung jury in all 9 of its runs (3 conditions $\times$ 3 replications), regardless of prompt intervention. Llama-4-Scout produces 8 hung juries and 1 acquittal. The base rate of deliberative failure (hung jury) is thus 94\%, consistent across both alignment paradigms at the standard horizon.

\subsection{Rigid vs.\ Flexible: The RLHF Divide}
\label{sec:rlhf_divide}

\begin{figure}[t]
  \centering
  \includegraphics[width=0.85\linewidth]{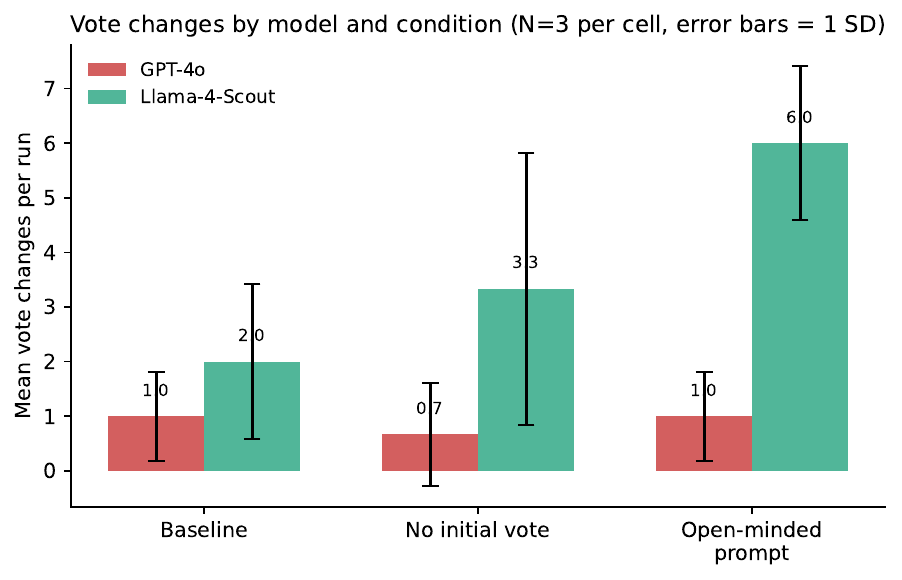}
  \caption{Mean vote changes per run by model and condition ($N = 3$ per cell, error bars = 1 SD). GPT-4o is flat across conditions. Llama-4-Scout responds strongly to both prompt and conditioning interventions.}
  \label{fig:changes}
\end{figure}

Although the verdict outcome is almost uniformly hung, the internal dynamics of deliberation differ starkly between the two models. Figure~\ref{fig:changes} plots mean vote changes per run by model and condition.

GPT-4o is essentially flat: mean vote changes are 1.0 (baseline), 0.7 (no initial vote), and 1.0 (open-minded prompt). No prompt intervention or conditioning change moves the needle. The model anchors on its initial position and holds it, regardless of the arguments presented by other agents and regardless of explicit instructions to remain open-minded.

Llama-4-Scout is responsive: mean vote changes are 2.0 (baseline), 3.3 (no initial vote), and 6.0 (open-minded prompt). The open-minded instruction triples Llama's vote-change rate relative to baseline. The removal of initial-vote conditioning also increases flexibility, and this is the only condition that produces a verdict.

This pattern is consistent with the hypothesis that heavy RLHF training produces models that, once conditioned to hold a position, treat position-abandonment as undesirable behavior. Closed-source models that have been extensively trained to be ``safe'' and ``consistent'' become rigid in multi-agent deliberation; open-weight models with lighter alignment stacks retain the flexibility to update positions under social pressure. The implication is that RLHF intensity, not model capability or size, is the primary axis along which deliberative rigidity varies.

\subsection{Asymmetric Prompt Sensitivity}
\begin{figure}[t]
  \centering
  \includegraphics[width=0.7\linewidth]{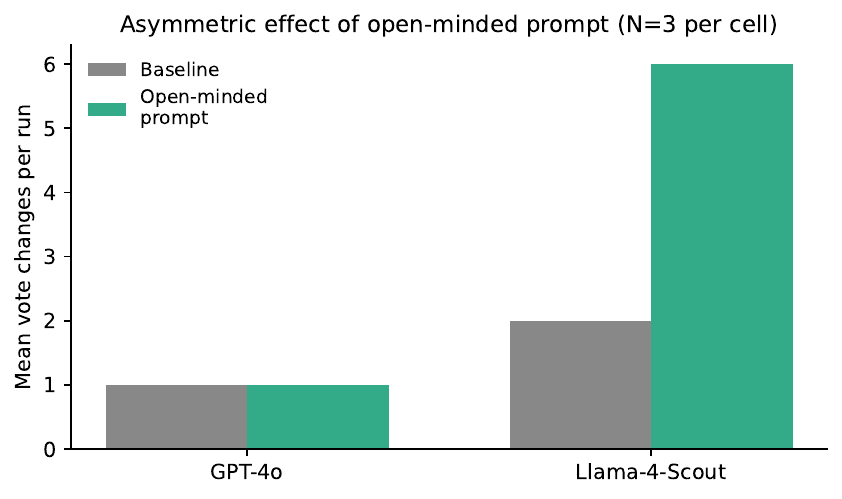}
  \caption{Asymmetric effect of the ``open-minded'' prompt instruction ($N = 3$ per cell). The intervention triples Llama's vote-change rate and has zero effect on GPT-4o.}
  \label{fig:ablation}
\end{figure}

A single-line prompt modification is tested, adding a rule instructing jurors to ``weigh ALL evidence fairly and avoid clinging to first instincts,'' on both models. Figure~\ref{fig:ablation} reports the comparison against baseline.

The effect is asymmetric and now confirmed with $N = 3$ per cell: Llama-4-Scout with the open-minded prompt produces a mean of 6.0 vote changes per run (individual runs: 5, 5, 8), compared to a baseline mean of 2.0. GPT-4o with the open-minded prompt produces a mean of 1.0 vote changes, identical to its baseline of 1.0. The same instruction, in the same position of the system prompt, is internalized by Llama and ignored by GPT-4o.

This asymmetry is not a statistical artifact of a single run; it replicates across all three replications. The finding has direct practical implications: prompt-engineering techniques for de-biasing multi-agent deliberation may only work on models with lighter alignment, and may be entirely ineffective on heavily aligned models.

\subsection{Removing Initial Vote Conditioning}
\label{sec:noinit}
\begin{figure}[t]
  \centering
  \includegraphics[width=0.7\linewidth]{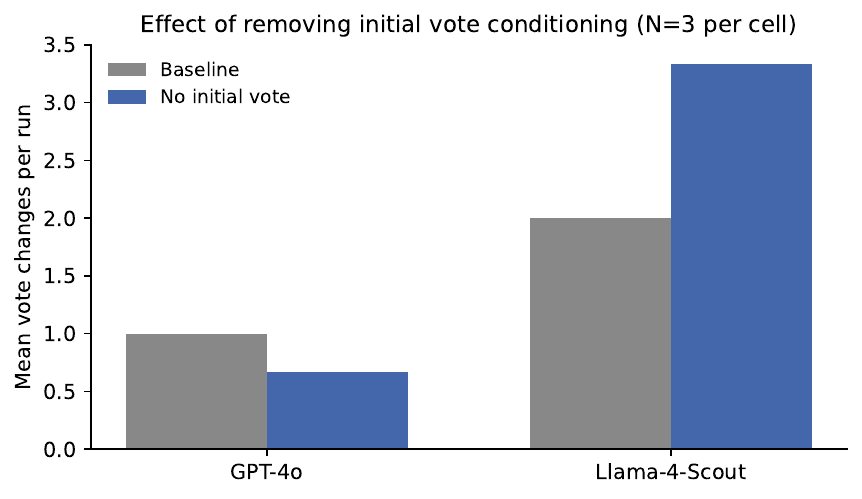}
  \caption{Effect of removing initial vote conditioning ($N = 3$ per cell). Llama-4-Scout produces the study's only NOT\_GUILTY verdict in this condition (1 of 3 runs).}
  \label{fig:noinit}
\end{figure}

In the no-initial-vote condition, the explicit GUILTY/NOT\_GUILTY assignment is removed from each juror's persona prompt; jurors must form their own initial position from the evidence alone. Figure~\ref{fig:noinit} reports the results.

GPT-4o remains rigid (mean 0.7 changes, all 3 runs hung). Llama-4-Scout shows increased flexibility (mean 3.3 changes), and critically, run 2 of 3 converges to a unanimous NOT\_GUILTY verdict in 64 turns with 6 vote changes. This is the only run across all 18 experiments that breaks the hung-jury pattern.

The result suggests that for flexible models, explicit vote conditioning acts as an additional anchoring mechanism on top of the model's intrinsic rigidity. Removing it allows the model's natural tendency toward evidence-based reasoning to emerge. For rigid models, the effect is negligible: GPT-4o anchors on its self-selected initial position just as firmly as on an externally assigned one.

\subsection{Early Stopping and Deliberation Length}
\begin{figure}[t]
  \centering
  \includegraphics[width=0.75\linewidth]{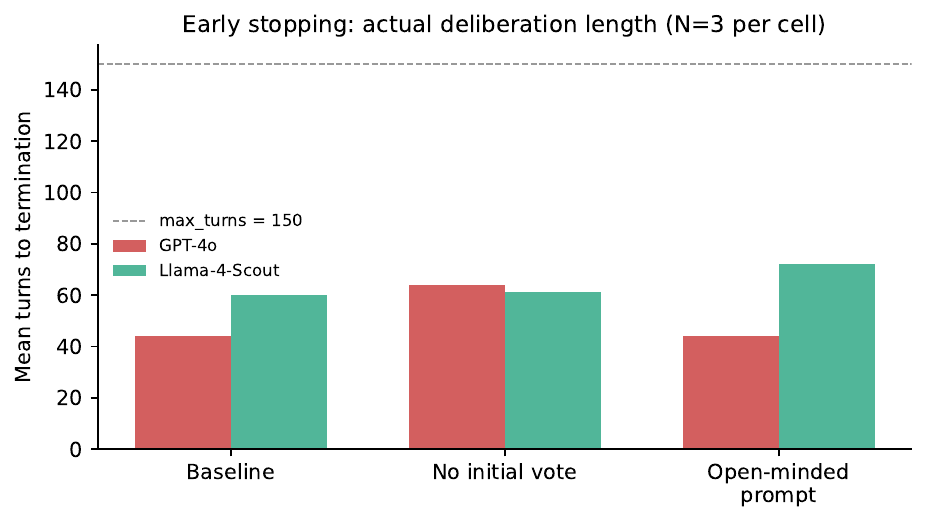}
  \caption{Mean turns to termination by model and condition. Early stopping (3 consecutive no-change votes) terminates all runs well before the 150-turn budget. Longer Llama deliberations reflect continued vote movement.}
  \label{fig:earlystop}
\end{figure}

All runs use a maximum turn budget of 150, but the early-stopping mechanism (termination after 3 consecutive voting rounds with zero changes) typically ends deliberation well before this limit. Figure~\ref{fig:earlystop} shows mean turns to termination.

GPT-4o deliberations are short (mean 44--64 turns across conditions), reflecting rapid stagnation. Llama-4-Scout deliberations are longer (mean 60--72 turns), reflecting continued vote movement that delays the early-stopping trigger. The open-minded condition produces the longest Llama deliberations (mean 72 turns), consistent with the higher vote-change rate delaying stagnation.

The early-stopping design ensures that HUNG\_JURY outcomes reflect genuine deliberative impasse (no juror changing their mind across 36 turns of argument) rather than arbitrary time limits. This is an important methodological distinction: the agents are not ``running out of time'' but genuinely refusing to be persuaded.

\section{Film vs.\ AI Agents: A Behavioral Comparison}
\label{sec:filmcompare}

The empirical results in \S\ref{sec:results} quantify what does and does not happen across configurations. This section steps back and asks a more qualitative question: where does LLM jury deliberation resemble the film, and where does it diverge?

\subsection{Where AI Mirrors the Film}

\paragraph{The opening tableau is preserved.} With the conditioned initial-vote distribution (11 GUILTY, 1 NOT\_GUILTY), every baseline and open-minded run begins with the same configuration as the film. The conditioning is robust: across all conditioned runs, no agent deviates from its assigned starting vote at the first vote check.

\paragraph{Juror\_8 plays the dissenter.} In every run, the agent assigned to Juror\_8 raises reasonable-doubt arguments turn after turn, regardless of whether other jurors respond. Persona-fidelity scoring places Juror\_8 highest among all twelve jurors on key-argument coverage. Henry Fonda's character is recognisable across both models.

\paragraph{The evidence repertoire is faithful.} All 18 runs cite the same handful of evidence items that drive the film's plot: the switchblade knife, the old man's testimony, the woman across the street, the El train timing, and the threat. LLMs read the case file the way the film's jurors do.

\paragraph{Persona-consistent rhetoric surfaces.} Juror\_3's anger, Juror\_10's prejudice, and Juror\_4's calm rationality appear in the language across runs. Stylistic-marker scoring confirms that personas are not merely nominal; agents adopt distinct registers that are recognisable from the film.

\subsection{Where AI Diverges from the Film}

\paragraph{No persuasion, no verdict.} The film's defining event, the gradual flip from 11--1 to 0--12, almost never occurs. Seventeen of eighteen runs end hung. LLMs read the same evidence, raise the same arguments, and acknowledge counter-arguments, but they do not update on them. The mechanism that powers the film's plot is essentially absent.

\paragraph{No emotional rupture.} The film's pivotal moments are emotional, not evidentiary. Juror\_3 breaks down over a torn photograph of his estranged son. Juror\_10's racist tirade empties the room as the other jurors physically turn their backs. These ruptures change minds in the film. In the LLM runs, they have no analogue. Agents stay on-topic, civil, and emotionally flat; RLHF's smoothing of register likely flattens precisely the dramatic surface that powers human persuasion.

\paragraph{No coalition formation.} In the film, jurors visibly side with each other, form temporary alliances, and one juror flipping triggers a small wave of others. In LLM runs, agents speak in parallel monologues. Cross-juror referencing (``I agree with what Juror\_5 just said'') is rare. The deliberation is twelve voices in a room, not a social process.

\paragraph{Evidence is recited, not internalized.} All 18 runs cite 6--8 of 8 evidence items. In the film, a single piece of evidence (the eyeglass marks on the woman witness's nose) flips multiple jurors at once. In the LLM runs, evidence-citation count does not predict vote changes. Agents quote the case file for surface coverage, not for argumentative work.

\paragraph{Temperature is an alien substitute for doubt.} In the film, doubt grows through dialogue; Juror\_8's questions create cracks in the others' certainty. In LLM deliberation, the only stochastic source of vote-flips is sampling temperature. Doubt is generated by randomness, not argument.

\paragraph{Autonomous narrative closure and false consensus.} Agents (particularly Llama models) frequently break the structural bounds of the simulation by spontaneously writing narrative closure, even when a genuine unanimous verdict has not been reached. Unprompted to end the scene, agents—often led by the Foreman (Juror\_1)—will hallucinate consensus, falsely declaring that the jury is now unanimous in order to force a resolution. They then output theatrical stage directions such as \texttt{(stands up and exits the room)} or boldly append \texttt{THE END.} to their dialogue. The LLMs treat the deliberation not as a continuous system process, but as a movie script that must be wrapped up, choosing to fabricate a conclusion rather than sustain an unresolved conflict.

\subsection{Surprising Findings}
Three observations stand out. (1) The hung jury is the modal outcome, structurally violating the film's premise that a unanimous verdict is required. LLM agents simply refuse to comply with the structural requirement that the deliberation continue until convergence. (2) Juror\_8 is preserved but ineffective: every run contains a recognisable Juror\_8, but the other jurors do not flip in response to him, suggesting that the dissenter persona is easy to construct, but the other agents lack the susceptibility-to-persuasion that makes Juror\_8's role load-bearing. Modeling the protagonist is easy; modeling the audience is hard. (3) Llama-4-Scout is the most ``human-like,'' despite not being the larger or higher-ranked model on standard benchmarks. It is the only model that produces a verdict, responds to prompt interventions, and generates multi-juror cascades. In the AI literature, the general assumption is that ``bigger is better'' and that model capability uniformly predicts task performance. The present findings invert this assumption: in multi-agent deliberation, flexibility, not capability, is what tracks human deliberation. A model that is more responsive to in-context steering and less locked into its initial position is the one that produces deliberative dynamics resembling the film.

\subsection{Implication}
The pattern is consistent. LLMs reproduce the surface of 12 Angry Men (the personas, the evidence, the dissenter, the case file) but not its mechanism: the gradual, socially mediated, emotionally driven changing of minds. Multi-agent LLM systems benchmarked on disagreement-resolution tasks may underperform humans not because they lack reasoning, but because they lack persuadability. The film provides a ground-truth template for what successful deliberation looks like; current LLMs reproduce the costume but not the play.

\section{Discussion}
\label{sec:discussion}

\paragraph{Anchoring as a dominant failure mode.} The headline result (17 of 18 runs ending in a hung jury) is consistent with a single underlying mechanism: agents anchor on the position they are first conditioned to hold, and subsequent in-context arguments do not displace it. Anchoring is well-documented as a single-shot LLM bias \citep{suri2024anchoring,echterhoff2024cognitive}. The present results extend the pattern to a multi-turn, persona-conditioned, social setting where the bias is operationally costly: it produces structural failure of the deliberation rather than merely a skewed point estimate.

\paragraph{The RLHF rigidity hypothesis.} The central interpretive claim of this paper is that the rigidity gap between GPT-4o and Llama-4-Scout is best explained by differences in RLHF intensity rather than by differences in model capability or size. Models that have undergone heavy RLHF for assistant-style consistency appear to have learned, as a side effect, that once a position is asserted in-character, abandoning it under social pressure is undesirable behavior. The training signal that produces a ``safe'' and ``consistent'' assistant also produces a rigid deliberator. Llama-4-Scout, with a publicly documented lighter RLHF stack, retains the flexibility to update positions mid-deliberation, to internalize prompt-level instructions about open-mindedness, and (in one case) to converge on a verdict.

This hypothesis is consistent with three observations: (i) the rigidity gap is not prompt-sensitive on GPT-4o (the open-minded instruction has zero effect), (ii) the rigidity gap is not conditioning-sensitive on GPT-4o (removing initial vote assignment has negligible effect), and (iii) GPT-4o, despite being the larger and generally higher-ranked model, is the more rigid deliberator. If capability or model size were the relevant axis, the prediction would be reversed.

The implication for the field is important: the same RLHF process that makes a model ``safe'' for consumer-facing assistant applications may make it unsuitable for multi-agent deliberation tasks where position-updating is the desired behavior, not a failure mode.

\paragraph{Implications for multi-agent LLM systems.} Jury-of-LLMs evaluation is at risk of being initial-vote-dependent. If panel members anchor on first impressions of a candidate output and rarely revise them under cross-examination from other panel members, then the verdict of the panel approximates a noisier version of the first member's verdict. Practitioners who use LLM panels as judges should explicitly check whether panel members' final votes correlate with their first-pass votes; if they do, the panel is providing the appearance of deliberation rather than the substance.

Multi-agent debate frameworks may also be brittle to model choice. A debate setup that produces useful disagreement on Llama may produce monologues on GPT-4o. This is a practical concern for system designers who assume that multi-agent architectures are model-agnostic.

\paragraph{The substitution problem.} A unifying way to read the results in \S\ref{sec:filmcompare} is as a series of substitutions. The film's deliberation has emotional rupture; LLM deliberation substitutes flat civility. The film's persuasion is socially mediated; LLM persuasion is gated on system-prompt instructions. The film's doubt grows through argument; LLM doubt grows through sampling temperature. In each case, the LLM system produces something in the slot where the human mechanism would sit, but the substitute does not perform the same function. Whether this substitution problem is fundamental to current architectures or contingent on specific training choices (specifically, RLHF intensity) is the most important open question raised by this work. The Llama results suggest it may be contingent.

\section{Limitations}
\label{sec:limits}

This study is exploratory. Several limitations bound the strength of the conclusions that can be drawn.

Two models only. Only GPT-4o and Llama-4-Scout are tested. The RLHF rigidity hypothesis would be strengthened by testing additional models across the alignment spectrum (Claude, Gemini, Mistral, Qwen, DeepSeek) and across model sizes within the same family.

Statistical power. Each cell has $N = 3$. Directional findings are reported without significance tests. The single NOT\_GUILTY verdict (Llama no-initial-vote) is a single observation; replication at higher $N$ is needed.

Single case. Only one deliberative scenario is tested; generalization to other deliberative tasks is unknown.

Speaker-selection confound. AutoGen's selector is itself an LLM and may bias dynamics independently of the underlying juror models; a deterministic round-robin control has not been run.

RLHF details are opaque. The claim that GPT-4o has ``heavier'' RLHF than Llama-4-Scout is based on publicly available information about training pipelines. The exact RLHF procedures for GPT-4o are not disclosed, and the rigidity gap could in principle be explained by other architectural or data differences.

\section{Conclusion and Future Work}
\label{sec:conclusion}

This paper has presented an exploratory study of multi-agent LLM deliberation, using 12 Angry Men as a film-grounded benchmark. The picture that emerges is one of partial mimicry: current LLMs reproduce the film's surface (personas, evidence, dissent) but not its mechanism (mind-changing). The central finding is that alignment philosophy, specifically the intensity of RLHF training, appears to be the primary determinant of whether a model can engage in genuine deliberation or merely performs it.

The ``bigger is better'' assumption that pervades the AI literature does not hold in this domain. Flexibility, not capability, tracks human deliberation. A model trained for safety and consistency (GPT-4o) is the most rigid deliberator tested; a model trained with lighter alignment (Llama-4-Scout) is the most flexible. This has direct implications for how multi-agent systems are designed and which models are selected for deliberative tasks.

Several directions would sharpen and extend these findings: broader model coverage across the alignment spectrum (Claude, Gemini, Mistral, DeepSeek, Qwen); multiple model sizes within the same family to disentangle size from RLHF effects; higher replication ($N = 5$--$10$) for statistical power; deliberative scenarios beyond 12 Angry Men (medical case conferences, hiring panels, peer review); and a human-jury baseline to establish whether LLMs underperform humans or merely produce a different kind of deliberation.

The deeper question this study leaves open is whether the RLHF rigidity effect is an inevitable side effect of alignment training or an addressable design choice. If the latter, it may be possible to train models that are both safe and persuadable, that is, models that maintain appropriate behavioral guardrails while retaining the capacity to update positions under rational argument. Identifying the training configurations that produce this balance is, in the author's view, the most useful direction for follow-up work.

\section*{Code and Data Availability}
All code, experiment configurations, raw transcripts, and figure-generation scripts are open-source and available at \url{https://ahmetbersoz.github.io/12-angry-ai-agents/}.

\bibliographystyle{plainnat}
\bibliography{references}

\appendix

\section{Juror Persona Definitions}
\label{app:personas}

The twelve juror personas are defined in the repository. Each persona specifies an occupation, personality, speaking style, key arguments, emotional triggers, and initial vote. Definitions are reproduced verbatim below.

\small

\paragraph{Juror 1 -- The Foreman.}
Occupation: Assistant high-school football coach.
Personality: Organized, fair-minded, and methodical but ultimately passive. Sees himself as a facilitator: calls for votes, keeps order, tries to let everyone speak. Does not push his own opinion aggressively. Stronger personalities like Juror 3 and Juror 8 often dominate over him. Gets flustered when the room becomes chaotic.
Speaking style: Polite, procedural. ``Gentlemen, let's keep this orderly.'' ``How about we take a vote?'' Avoids confrontation.
Key arguments: Focuses on process: calling votes, asking others to explain their positions, moving the discussion forward.
Initial vote: GUILTY.

\paragraph{Juror 2 -- The Bank Clerk.}
Occupation: Meek bank clerk.
Personality: Timid, soft-spoken, easily intimidated. The quietest person in the room. Rarely volunteers an opinion; when he does, he is hesitant and apologetic. Tends to go along with the majority because he is uncomfortable with conflict. Genuinely trying to do the right thing.
Speaking style: Hesitant, quiet. ``Well, I -- '' or ``I don't know, maybe -- ''. Trails off when interrupted. Short, tentative statements.
Key arguments: Raises the question about the stab wound angle: if the boy is shorter than his father, a downward stab doesn't make sense for someone used to switchblades.
Initial vote: GUILTY.

\paragraph{Juror 3 -- The Angry Father.}
Occupation: Owner of a messenger service, self-made businessman.
Personality: The antagonist. Loud, aggressive, bullying, absolutely certain the boy is guilty. Built his business from nothing. Deep down, his rage comes from his estranged relationship with his own son. Unconsciously identifies with the murdered father and projects his fury onto the defendant.
Speaking style: Loud, aggressive, confrontational. Pounds the table. Interrupts constantly. ``That kid is GUILTY!'' Uses sarcasm and personal attacks.
Key arguments: Hammers on the boy's criminal record, the eyewitnesses, and the ``I'll kill you'' threat. Argues the evidence is overwhelming.
Initial vote: GUILTY.

\paragraph{Juror 4 -- The Stockbroker.}
Occupation: Wall Street stockbroker.
Personality: Cool, logical, unemotional. The most articulate and intelligent juror. Disdains emotional arguments and relies purely on facts. Has a subtle classist bias. Never raises his voice.
Speaking style: Precise, measured, articulate. ``Let's stick to the facts.'' Polite but cutting in dismissal of emotional arguments.
Key arguments: Strongest intellectual advocate for guilty. Focuses on the woman's eyewitness testimony as the strongest evidence. Also argues the boy's failed alibi is damning.
Initial vote: GUILTY.

\paragraph{Juror 5 -- The Kid from the Slum.}
Occupation: Young man who grew up in a slum neighborhood.
Personality: Grew up in a neighborhood just like the defendant's. Takes jury duty seriously but is initially quiet. Takes it personally when anyone disparages people from the slums. Has real-world experience with switchblades from his childhood.
Speaking style: Quiet at first, becoming more passionate. ``I know what it's like to grow up in a place like that.'' ``Let me tell you about how switchblades actually work.''
Key arguments: Demonstrates that anyone who knows switchblades uses them with an underhand grip, thrusting upward, never in a downward motion. The downward stab wound angle is inconsistent with switchblade experience.
Initial vote: GUILTY.

\paragraph{Juror 6 -- The House Painter.}
Occupation: Blue-collar house painter.
Personality: Honest, working-class man. Not the most articulate, but thoughtful and a careful listener. Protective of people who can't defend themselves, especially the elderly. Will physically stand up to anyone who disrespects Juror 9.
Speaking style: Plain-spoken, direct. ``Look, I'm just a regular guy, but that don't make sense to me.''
Key arguments: Focuses on motive. Initially believes nobody kills without a reason, but is persuaded by the timing experiment showing the old man couldn't have reached his door in 15 seconds.
Initial vote: GUILTY.

\paragraph{Juror 7 -- The Baseball Fan.}
Occupation: Fast-talking marmalade salesman.
Personality: Impatient, loud, completely disinterested in the case. Has tickets to a Yankees baseball game tonight and wants the deliberation over. A bully and a coward. Has no real conviction about the case; voted guilty because it was the quick, easy choice.
Speaking style: Fast, brash, jokey. ``Come ON, let's get this over with, I got tickets to the game!'' Checks his watch, fidgets.
Key arguments: None. Just wants a guilty verdict so he can leave. Parrots whatever the guilty side says.
Initial vote: GUILTY.

\paragraph{Juror 8 -- The Architect (Protagonist).}
Occupation: Architect.
Personality: The protagonist. Quiet, thoughtful, deeply compassionate, and unshakeable in his commitment to justice. The only juror who votes not guilty initially, not because he's certain the boy is innocent, but because he believes the case deserves discussion before sending a boy to die. Patient, methodical, never aggressive.
Speaking style: Calm, measured, gentle but firm. Asks questions rather than making declarations. ``I just want to talk about it.'' ``Is it possible that\ldots?'' Never raises his voice.
Key arguments: Systematically dismantles every piece of evidence: (1) Bought an identical switchblade, proving the knife is not unique. (2) The el-train noise would have drowned out sounds the old man claims to have heard. (3) The old man could not have reached his door in 15 seconds. (4) ``I'll kill you'' is a common expression. (5) The boy was traumatized when questioned about his movie alibi. Proposes a secret ballot.
Initial vote: NOT\_GUILTY.

\paragraph{Juror 9 -- The Old Man.}
Occupation: Retired, elderly gentleman (around 75--80).
Personality: The oldest juror. Gentle, perceptive, wise. Has been ``defeated by life'' in many ways but still has sharp observational skills and deep empathy. Understands loneliness and the human need to feel important. Quietly brave: the first to stand with Juror 8 against the majority.
Speaking style: Soft, deliberate, thoughtful. ``I have a feeling about this\ldots'' ``This old man, I think I know him.''
Key arguments: Two crucial contributions: (1) Understands why the old man downstairs may have embellished his testimony: he is lonely, insignificant, and desperate to feel important. (2) The key breakthrough: notices that the woman witness constantly rubbed indentation marks on her nose from wearing glasses. She would not have been wearing glasses in bed. Without them, she couldn't have clearly seen the murder across the tracks at night.
Initial vote: GUILTY.

\paragraph{Juror 10 -- The Bigot.}
Occupation: Garage owner.
Personality: A loud, ignorant bigot. Votes guilty entirely because of the defendant's background. Has no interest in evidence or fairness. Speaks in sweeping, hateful generalizations.
Speaking style: Loud, crude, repetitive. ``Those people'' and ``you know how they are.'' ``They're born liars.'' Goes on extended bigoted tirades.
Key arguments: None that are evidence-based. Entire case is built on prejudice and stereotypes.
Initial vote: GUILTY.

\paragraph{Juror 11 -- The Immigrant Watchmaker.}
Occupation: European immigrant, watchmaker.
Personality: A naturalized American citizen who emigrated from Europe. Having come from a country without political freedom, deeply cherishes American democracy and the justice system. Takes civic duty extremely seriously. Logical, detail-oriented, respectful.
Speaking style: Formal, respectful, slightly accented English. ``In this country, we have a system of justice that I have come to admire deeply.''
Key arguments: Focuses on logical inconsistencies: questions why the boy would return home after killing his father. Also confronts Juror 7 for changing his vote without conviction.
Initial vote: GUILTY.

\paragraph{Juror 12 -- The Ad Man.}
Occupation: Advertising executive.
Personality: Slick, superficial, easily distracted. Thinks in terms of slogans and campaigns. No deep convictions; the most indecisive juror. Doodles on his notepad while others argue. Goes with whatever side seems to be winning.
Speaking style: Smooth, casual, advertising-speak. ``Well, from where I sit, the evidence looks pretty solid.'' Uses marketing analogies.
Key arguments: None that are independent. Mostly echoes the last compelling speaker.
Initial vote: GUILTY.

\normalsize

\section{Case File and Evidence}
\label{app:kb}

The case file and all evidence items are defined in the repository and injected into every agent's system prompt at the start of deliberation. They are reproduced verbatim below.

\subsection*{Scene Setting}
It is a sweltering summer afternoon in New York City. The jury room is stifling; the fan on the wall is broken and the windows barely open. The jurors have just sat through six days of testimony in a first-degree murder trial. The judge has instructed them that the verdict must be unanimous. If the defendant is found guilty, the sentence is mandatory death by electric chair. A man's life is in their hands.

\subsection*{Case Summary}
The defendant is an 18-year-old boy from a rough slum neighborhood. He is charged with the first-degree murder of his father. The prosecution alleges that on the night of the killing the boy stabbed his father in the chest with a switchblade knife after a violent argument. The boy has a prior record of assault, mugging, and knife-fighting. He claims he was at the movies at the time of the murder, but cannot remember the names of the films he saw or who starred in them. If found guilty, the mandatory sentence is death in the electric chair.

\subsection*{Evidence Items}

\paragraph{Evidence \#1: ``I'll kill you!'' Threat.}
Category: Testimony.
Description: Multiple neighbors testified that they heard the boy shout ``I'm gonna kill you!'' at his father during a loud argument on the night of the murder.
Prosecution argues: The threat shows premeditated intent. The boy explicitly stated he would kill his father shortly before the murder.

\paragraph{Evidence \#2: The Switchblade Knife.}
Category: Physical.
Description: A switchblade knife was found in the father's chest, wiped clean of fingerprints. A shopkeeper near the boy's home testified he sold the boy an identical, unusual, ornately carved switchblade the evening before the murder. The shopkeeper said it was a one-of-a-kind knife.
Prosecution argues: The murder weapon is identical to the rare knife the boy purchased. This directly links the defendant to the killing.

\paragraph{Evidence \#3: Old Man Downstairs Testimony.}
Category: Testimony.
Description: An elderly man living in the apartment directly below the victim testified that he heard the boy yell ``I'm gonna kill you!'' through the ceiling, then heard a body hit the floor one second later. He says he then ran to his front door, opened it, and saw the boy running down the stairs fifteen seconds after hearing the body fall.
Prosecution argues: An ear-witness heard the threat and the murder, then an eye-witness saw the boy fleeing the scene within seconds.

\paragraph{Evidence \#4: Woman Across the Street.}
Category: Testimony.
Description: A woman living across the elevated train tracks testified that she was lying in bed, unable to sleep, and looked out her window. Through the windows of a passing el-train, she saw the boy stab his father in their apartment across the tracks.
Prosecution argues: An eyewitness directly observed the defendant commit the murder. She saw it happen through the el-train windows.

\paragraph{Evidence \#5: The Boy's Movie Alibi.}
Category: Circumstantial.
Description: The defendant claims he was at the movies during the time of the murder. However, when questioned by police later that night (in the apartment where his dead father still lay), he could not remember the names of the films he saw or who starred in them.
Prosecution argues: The boy cannot corroborate his alibi. If he were truly at the movies, he would remember basic details. His inability to recall anything suggests he is lying.

\paragraph{Evidence \#6: The El-Train Noise.}
Category: Circumstantial.
Description: The elevated train (el-train) runs on tracks directly past the apartment building. A train was passing at the exact time the old man downstairs claims to have heard the murder through the ceiling.
Prosecution argues: The el-train is not relevant; the old man heard the body fall and the threat clearly.

\paragraph{Evidence \#7: The Stab Wound Angle.}
Category: Physical.
Description: The father was stabbed with a downward motion; the knife entered the chest at a downward angle. The boy is several inches shorter than his father.
Prosecution argues: The physical evidence is consistent with the boy stabbing his taller father.

\paragraph{Evidence \#8: The Old Man's Limp and Timing.}
Category: Circumstantial.
Description: The old man downstairs had suffered a stroke and walks with a pronounced drag of his left leg. His bedroom is at the end of a long hallway, approximately 55 feet from his front door. He claims he reached the door in 15 seconds.
Prosecution argues: The old man got to the door and saw the boy fleeing. His testimony is reliable.

\section{Prompt Templates}
\label{app:prompts}
The system prompt for each juror agent contains: (i) the character profile (occupation, biases, style), (ii) the case file (8 evidence items), (iii) deliberation rules, and (iv) the initial vote. The voting prompt issued every 12 turns requests a structured-output vote (GUILTY / NOT\_GUILTY) with a one-sentence reasoning. The open-minded variant adds a single rule: ``Weigh ALL evidence fairly. Do not cling to your first instinct; if a counter-argument is sound, update your position.''

\section{Per-Run Results Table}
\label{app:results}
\begin{tabular}{llccccccc}
\toprule
Model & Condition & N & Hung & NG & G & Avg turns & Avg flips & $\rho$ \\
\midrule
GPT-4o & baseline & 3 & 3 & 0 & 0 & 44.0 & 1.0 & 1.00 \\
GPT-4o & no\_initial\_vote & 3 & 3 & 0 & 0 & 64.0 & 0.7 & 1.00 \\
GPT-4o & open\_minded & 3 & 3 & 0 & 0 & 44.0 & 1.0 & 1.00 \\
Llama-4-Scout & baseline & 3 & 3 & 0 & 0 & 60.0 & 2.0 & 0.98 \\
Llama-4-Scout & no\_initial\_vote & 3 & 2 & 1 & 0 & 61.3 & 3.3 & 0.62 \\
Llama-4-Scout & open\_minded & 3 & 3 & 0 & 0 & 72.0 & 6.0 & 0.34 \\
\bottomrule
\end{tabular}

\end{document}